\documentclass{article}
\usepackage{spconf}
\usepackage{ifpdf}
\usepackage{cite}
\usepackage{graphicx}
\usepackage{amsmath}
\usepackage{algorithmic}
\usepackage{array}
\usepackage{blindtext}
\usepackage{epstopdf}
\usepackage{bm}
\usepackage{enumerate}
\usepackage{amsbsy}
\usepackage{amssymb}
\usepackage{amsthm}
\usepackage{amscd}
\usepackage{subfigure}
\usepackage{color}
\usepackage{booktabs}
\usepackage{multirow}
\usepackage{hhline}
\usepackage{makecell}
\usepackage{bbold}
\usepackage{url}
\usepackage{tabularx}
\usepackage{cellspace}
\usepackage{boldline}
\usepackage{balance}
\usepackage[]{algorithm2e}
\usepackage{dsfont}

\def \bd           {\boldsymbol{d}}

\def \br           {\boldsymbol{r}}
\def \bs           {\boldsymbol{s}}
\def \bt           {\boldsymbol{t}}

\def \bv           {\boldsymbol{v}}

\def \bx           {\boldsymbol{x}}
\def \by           {\boldsymbol{y}}
\def \bz           {\boldsymbol{z}}

\def \bR           {\boldsymbol{R}}

\def \bdelta       {\boldsymbol{\delta}}

\def \btau         {\boldsymbol{\tau}}

\def \bUpsilon     {\boldsymbol{\Upsilon}}

\def \bPhi         {\boldsymbol{\Phi}}

\def \calD         {\mathcal{D}}

\def \calG         {\mathcal{G}}

\def \calN         {\mathcal{N}}

\def \calR         {\mathcal{R}}

\def \bzero        {\boldsymbol{0}}
\def \bone         {\boldsymbol{1}}

\def \expecE       {\mathds{E}}

\def \realR        {\mathds{R}}

\def \cdiag        {\mathrm{Diag}}

\def \st           {\mathrm{s.t.}}

\def \sign         {\mathrm{sign}}
\def \relu         {\mathrm{ReLU}}

\title{Deep One-bit Compressive Autoencoding}

\name{Shahin~Khobahi$^*$, Arindam~Bose and Mojtaba~Soltanalian \thanks{$^*$ Corresponding author (e-mail: \textit{skhoba2@uic.edu}). This work was supported in part by U.S. National Science Foundation Grants CCF-1704401.}}
\address{ECE Department, University of Illinois at Chicago, Chicago, IL 60607}

\begin{document}
	
	\maketitle
	\begin{abstract}
		Parameterized mathematical models play a central role in understanding and design of complex
		information systems. However, they often cannot take into
		account the intricate interactions innate to such systems. On the contrary, purely data-driven approaches do not need explicit mathematical models for data generation and have a wider applicability at the cost of interpretability. In this paper, we consider the design of a one-bit compressive autoencoder, and propose a novel hybrid model-based and data-driven methodology that allows us to not only design the sensing matrix for  one-bit data acquisition, but also allows for learning the latent-parameters of an iterative optimization algorithm specifically designed for the problem of one-bit sparse signal recovery. Our results demonstrate a significant improvement compared to state-of-the-art model-based algorithms.
	\end{abstract}
	
	\begin{keywords}
		Compressive sensing, low-resolution signal processing, deep unfolding, deep neural networks, autoencoders 
	\end{keywords}

	\vspace{-5mm}
	\section{Introduction} \label{sec:intro}
In the past decade, compressive sensing (CS) has shown significant potential in enhancing signal sensing and recovery performance with simpler hardware resources, and thus, has attracted noteworthy attention among researchers. CS is a method of signal acquisition which ensures the exact or almost exact reconstruction of certain classes of signals using far less number of samples than that is needed in Nyquist sampling \cite{4472240}---where the signals are typically reconstructed by finding the sparsest solution of an under-determined system of equations using various available means.

Note that in a practical settings, each measurement needs to be digitized into finite-precision values for further processing and storage purposes, which inevitably introduces a quantization error.
This error is generally dealt with as measurement noise possessing limited energy; an approach that does not perform well in extreme cases.
One-bit CS is one such extreme case where the quantizer is a simple sign comparator and each measurement is represented using only one bit, i.e., $+1$ or $-1$ \cite{ 4558487, 5955138, 6418031, 6404739, 6178284, zhang2014efficient}. 
One-bit quantizers are not only low-cost and low-power hardware components, but also much faster than traditional scalar quantizers, accompanied by great reduction in the complexity of hardware implementation.
Several algorithms have been introduced in the literature for efficient reconstruction of sparse signals in one-bit CS scenarios (e.g., see \cite{4558487, 5955138, 6418031, 6404739, 6178284,zhang2014efficient} and the references therein).
\nocite{khobahi2018signal,ameri2019one,roth2018comparison,zahabi2019one}
\vspace{-5mm}
\subsection{Relevant Prior Art}
The current one-bit CS recovery algorithms generally exploit the \textit{consistency principle}, which assumes that the element-wise product of the sparse signal and the corresponding measurement is always positive \cite{4558487}.
In \cite{4558487}, the authors introduce a regularizer based recovery algorithm called \textit{renormalized fixed point iteration} (RFPI) using a convex barrier function as a regularizer for the consistency principle.
We discuss the formulation of RFPI in Section \ref{sec:prob} in details.
Another such reconstruction algorithm can be found in \cite{5955138}, referred to as \textit{restricted step shrinkage} (RSS), for which a nonlinear barrier function is used as the regularizer. 
Compared to RFPI algorithm, RSS has three important advantages: provable convergence, improved consistency, and feasible performance~\cite{Li2018}.
Ref. \cite{6418031} introduces a penalty-based robust recovery algorithm, called \textit{binary iterative hard thresholding} (BIHT), in order to enforce the consistency principle.
Contrary to RFPI algorithm, BIHT needs the sparsity level of the signal as input.
Both RFPI and BIHT, however, perform very poorly in the presence of measurement noise, when \textit{bit flips} occur. 
In \cite{6404739} and \cite{6178284}, authors proposed modified versions of RFPI and BIHT, referred to as \textit{noise-adaptive renormalized fixed point iteration} (NARFPI) and \textit{adaptive outlier pursuit with sign flips} (AOP-f), that are robust against bit flips in the measurement vector. 

We note that model-based algorithms, such as the ones discussed above, often do not take into account the intricate interactions and the latent-parameters innate to complex signal processing systems. Therefore, there has recently been a high demand for developing effective real-time signal processing algorithms that use the data to achieve improved performance \cite{7780424, ILIADIS20189, 7447163}. In particular, the data-driven approaches relying on deep neural architectures such as convolutional neural networks \cite{7780424}, deep fully connected networks \cite{ILIADIS20189}, and stacked denoising autoencoders \cite{7447163}, have been studied for sparse signal recovery in generic quantized CS setting. Such data-driven approaches do not need explicit mathematical models for data generation and have a wider applicability. On the other hand, they lack the interpretability and trustability that comes with model-based signal processing techniques. 
The advantages associated with both model-based and data-driven methods show the need for developing approaches that enjoy the benefits of both frameworks  \cite{hershey2014deep,8683876}.

In this paper, we bridge the gap between the data-driven and model-based approaches in the one-bit CS area and propose a specialized, yet hybrid, methodology for the purpose of sparse signal recovery from one-bit measurements.
	
	\section{Problem Formulation}\label{sec:prob}

In a one-bit CS scenario, the dynamics of the data acquisition process (i.e., the encoder module) can be formulated as:
\begin{eqnarray}
\label{eq:7}
\text{Encoder Module: }\quad\quad\br = \sign(\bPhi\bx),
\end{eqnarray}
where $\bPhi^{m\times n}$ denotes the sensing matrix, and $\bx\in\mathds{R}^n$ is assumed to be a $K$-sparse signal. Having the one-bit measurements of the form \eqref{eq:7}, one can pose the problem of sparse signal recovery from one-bit measurements $\br$ by solving the following non-convex program:
	\begin{eqnarray}
	\underset{\bx}{\min}\; \|\bx\|_0,\label{eq:8} \;\st \;\br = \sign(\bPhi\bx), \label{eq:9}
	\end{eqnarray}
where the constraint in \eqref{eq:9} is imposed to ensure a consistent reconstruction from the one-bit information. Further note that, the one-bit measurement consistency principle can be equivalently expressed as $\bR \bPhi \bx \succeq \bzero$, where $\bR = \cdiag(\br)$ and $\succeq$ is an element-wise matrix inequality operator. Inspired by the CS literature, the above non-convex optimization problem can be further reformulated as a non-convex $\ell_1$-minimization program on the unit sphere:
	\begin{eqnarray}
		\underset{\bx}{\min} \;\|\bx\|_1,\;
		\st\;\bR \bPhi \bx \succeq \bzero,\;
		\; \|\bx\|_2 = 1,\label{eq:12}
	\end{eqnarray}
where the $\ell_1$-norm acts as a sparsity inducing function. The intuition behind finding the sparsest signal on the $\ell_2$ unit-sphere (i.e., fixing the energy of the recovered signal) is two-fold. First, it significantly reduces the feasible set of the optimization problem, and second, it avoids the the trivial solution of $\hat{\bx} = \bzero$. There exists an extensive body of research on solving the above non-convex optimization problem (e.g., see \cite{4558487, 5955138, 6404739, Plan2013, 6638799, zhang2014efficient}, and the references therein). The most notable methods utilize a regularization term $\calR(\bs)$ to enforce the consistency principle as a penalty term for the $\ell_2$-objective function, viz.
	\begin{eqnarray}
	\label{eq:6}
	\hat{\bx} = \underset{\bx}{\arg\min}\;  \|\bx\|_1  + \alpha\calR(\bR\bPhi\bx),\;
	\st	\;\|\bx\|_2 = 1,
	\end{eqnarray}
where $\alpha>0$ is the penalty factor. In this paper, we build upon the work done in \cite{4558487} and its proposed RFPI algorithm as a base-line to design the decoder function of the proposed one-bit compressive autoencoder (AE). In particular, we unfold the iterations of a renormalized fixed-point algorithm onto the layers of a neural network in a fashion that each layer of the proposed deep architecture mimics the behavior of one iteration of the base-line algorithm. Next, we perform an end-to-end learning approach by utilizing the back-propagation method to tune the parameters of both the decoder (i.e., parameters of the RFP iterations) and the encoder (i.e., the sensing matrix $\bPhi$) functions of the proposed compressive AE.

Let $\by = \bR\bPhi\bx$. In order to enforce the first constraint in~\eqref{eq:6}, the RFPI algorithm utilizes the regularization term
 $\rho(\by) \triangleq \max\{-\by,\bzero\}$. Note that the function $\rho(\cdot)$ can be expressed in terms of the well-known $\relu$ function extensively used by the deep learning research community, i.e. $\rho(\by) = \relu(-\by)$. The RFPI algorithm is a first-order optimization method (gradient-based) that operates as follows: Given an initial point $\bx_0$ on the unit-sphere, the gradient step-size $\delta$ and a shrinkage thresholds $\alpha$, at each iteration $i$, the estimated signal $\bx_{i}$ is obtained using the following update steps:
\begin{subequations}
	\begin{align}
		\label{eq:16}
		&\bd_i = \underset{\bx}{\nabla}\calR(\by)\bigr\rvert_{\bx = \bx_{i-1}} = -\left(\bR\bPhi\right)^T\rho\left(\bR\bPhi\bx_{i-1}\right),\\
		\label{eq:17}
		&\bt_i = \left(1 + \delta\bd_i^T\bx_{i-1}\right)\bx_{i-1} - \delta\bd_i,\\
		\label{eq:18}
		&\bv_i = \sign\left(\bt_i\right)\odot\relu\left(|\bt_i| - (\delta/\alpha)\bone\right),\\
		\label{eq:19}
		&\bx_i = \frac{\bv_i}{\|\bv_i\|_2}.
	\end{align}
\end{subequations}
After the descent in \eqref{eq:16}-\eqref{eq:17}, the update step in \eqref{eq:18} corresponds to a shrinkage step. More precisely, any element of the vector $\bt_i$ that is below the threshold $\delta/\alpha$ will be pulled down to zero (leading to enhanced sparsity). Finally, the algorithm projects the obtained vector $\bv_i$ on the unit sphere to produce the latest estimation of the signal.

While effective in signal reconstruction, 
there exist several drawbacks in using the RFPI method. For instance, it is required to use the algorithm on several problem instances, while increasing the value of the penalty factor $\alpha$ at each outer iteration of the algorithm, and to use the previously obtained solution as the initial point for tackling the recovery problem for any new problem instance. Moreover, it is not straight-forward how to choose the fixed step-size and the shrinkage threshold, that may depend on the latent-parameters in the information system. In fact, it is evident that by carefully tuning the step-sizes and the shrinkage threshold $\tau = \delta/\alpha$, one can significantly boost the performance of the algorithm, and further alleviate the mentioned drawbacks of this method. In the next section, we show how this tuning can be done by learning from the data. In particular, we slightly modify and over-parameterize the above updating steps of the RFPI algorithm and unfold them onto the layers of a deep neural network, and define a decoder function based on the unfolded iterations, and seek to jointly learn the parameters of the proposed~AE. 

	\section{The Proposed One-Bit Compressive  AutoEncoder}\label{sec:method}
	\vspace{-.2cm}
We pursue the design of a novel \emph{one-bit compressive sensing-based autoencoder} architecture that allows us to jointly design the parameters of both the encoder and the decoder module when one-bit quantizers are employed in the data acquisition process (i.e., the encoding module) for a $K$-sparse input signal $\bx\in\realR^n$. Briefly speaking, an AE is a generative model comprised of an encoder and a decoder module that are sequentially connected together. The purpose of an AE is to learn an abstract representation of the input data, while providing a powerful data reconstruction system through the decoder module. The input to such system is a set of signals following a certain distribution, i.e. $\bx\sim\calD(\bx)$, and the output is the recovered signal from the decoder module $\hat{\bx}$. Hence, the goal is to jointly learn an abstract representation of the underlying distribution of the signals through the encoder module, and simultaneously, learning a decoder module allowing for reconstruction of the compressed signals from the obtained abstract representations. Therefore, an AE can be defined by two functionals: \emph{i}) an encoder function $f^{\text{Encoder}}_{\bUpsilon_1}: \realR^n \mapsto \realR^m$, parameterized on a set of variables $\bUpsilon_1$ that maps the input signal into a new vector space, and \emph{ii}) a decoder function $f^{\text{Decoder}}_{\bUpsilon_2}: \realR^m \mapsto \realR^n$ parameterized on $\bUpsilon_2$, which maps the output of the encoder module back into the original signal space. Hence, the governing dynamics of a general AE can be expressed as $\hat{\bx} =  f^{\text{Decoder}}_{\bUpsilon_2} \circ f^{\text{Encoder}}_{\bUpsilon_1}(\bx)$,
%
where $\hat{\bx}$ denotes the reconstructed signal.

	\begin{figure*}
		\centering
		\subfigure[]{
			\centering
			\includegraphics[width=0.4\textwidth]{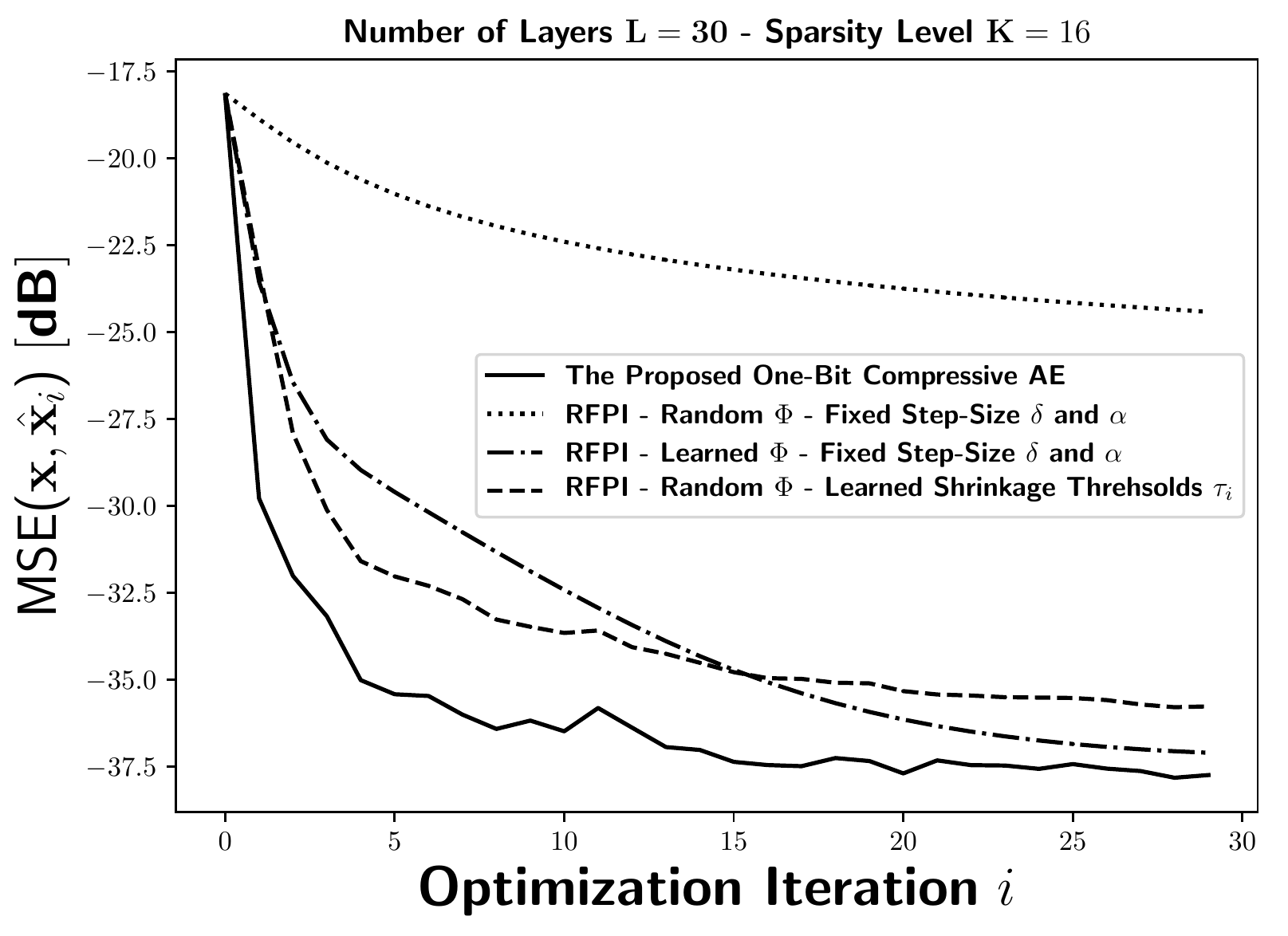}
		} \qquad\qquad
		\subfigure[]{
			\centering
			\includegraphics[width=0.4\textwidth]{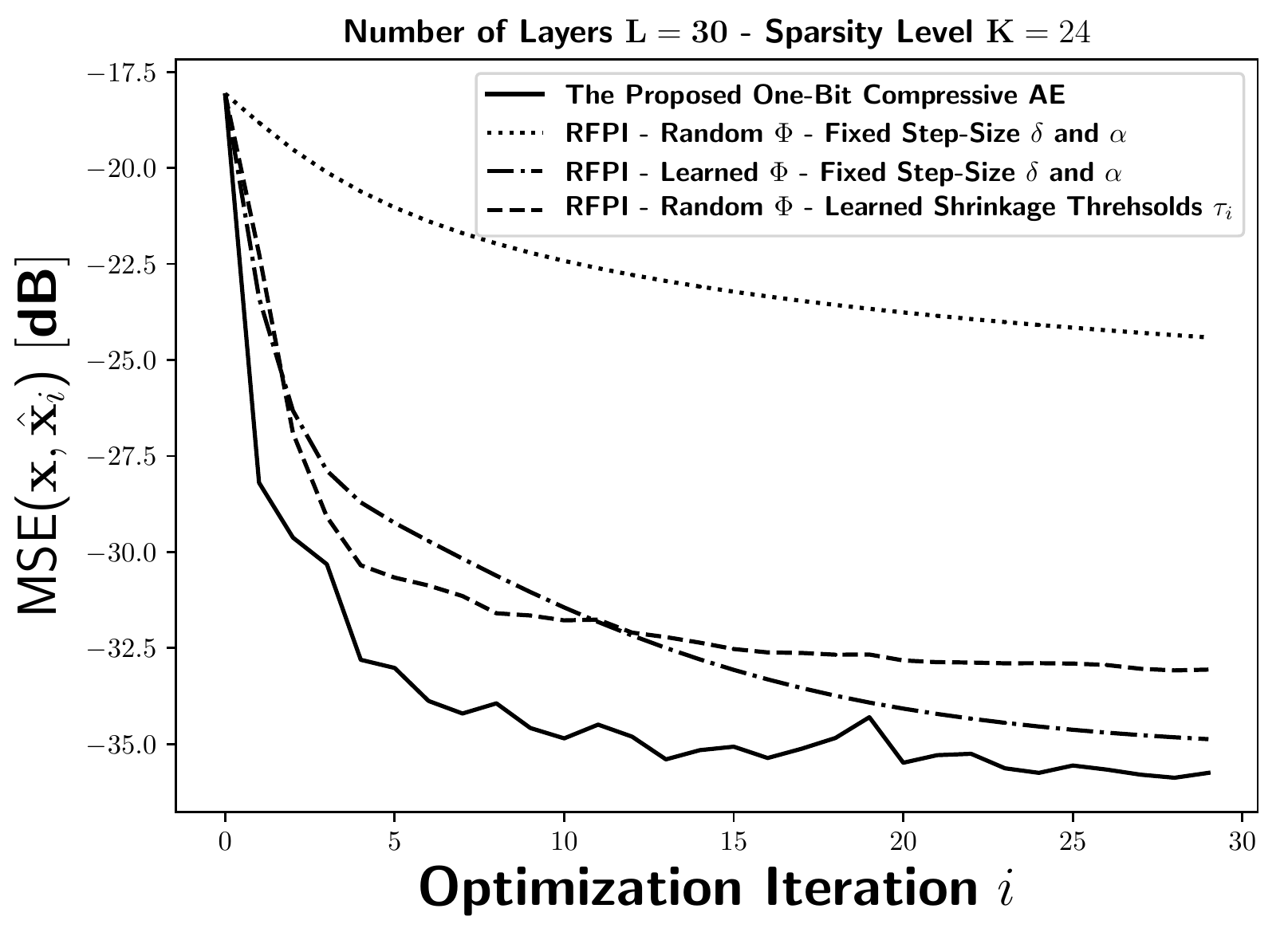}
		}
		\vspace{-.2cm}
		\caption{\small The performance of the proposed method compared to the RFPI algorithm for sparsity levels: (a) $K=16$, and (b) $K=24$.}
		\label{fig:2}
		\vspace{-.3cm}
	\end{figure*}

In light of the above, we seek to interpret a one-bit CS system as an AE module facilitating not only the design of the sensing matrix $\bPhi$ that best captures the information of a $K$-sparse signal when one-bit quantizers are employed, but also to learn the parameters of an iterative optimization algorithm specifically designed for the task of signal recovery. To this end, we modify and unfold the iterations of the form \eqref{eq:16}-\eqref{eq:19} onto the layers of a deep neural network and later use the deep-learning tools to tune the parameters of the proposed one-bit compressive AE. In particular, we define the encoder module of the proposed AE as follows:
\begin{eqnarray}
\label{eq:20}
f^{\text{Encoder}}_{\bUpsilon_1}(\bx) = \tilde{\sign}(\bPhi\bx),
\end{eqnarray}
where $\bUpsilon_1=\{\bPhi\}$ denotes the set of parameters of the encoder function, and  $\tilde{\sign}(\bx) = \tanh(c\cdot\bx)$, for some choice of $c>0$ ($c$ was set to $100$ in numerical investigations). Note that we replaced the original $\sign$ function with a smooth approximation of it based on the hyperbolic tangent function. The reason for such a replacement is that the $\sign$ function is not continuous and its gradient is zero everywhere except at the origin, and hence, the use of it would cripple any stochastic gradient-based optimization method (used in backpropagation method in deep learning). 
As for the decoder function, define $g_{\phi_i}: \realR^m \mapsto \realR^n$ as follows:
\begin{subequations}
	\label{eq:10}
\begin{align}
\label{eq:10a}
	&g_{\phi_i}(\bz;\bPhi,\bR) = \frac{\bv}{\|\bv\|_2},\quad \text{with}\\
\label{eq:10b}
	&\bv = \tilde{\sign}\left(\bt\right)\odot\relu\left(|\bt| - \btau_i\right),\\
	\label{eq:10c}
	&\bt = \left(1 + \delta_i\bd^T\bz\right)\bz - \delta_i\bd,\\
	\label{eq:10d}
	&\bd = -\left(\bR\bPhi\right)^T\rho\left(\bR\bPhi\bz\right),
\end{align}
\end{subequations}
where $\phi_i = \{\btau_i, \delta_i\}$ represents the parameters of the function $g_{\phi_i}$, and $\btau_i\in\realR^{n}$ denotes the sparsity inducing thresholds vector. Next, we define the proposed composite decoder function as follows:
\begin{eqnarray}
	f^{\text{Decoder}}_{\bUpsilon_2}(\bz_0) = g_{\phi_{L-1}}\circ g_{\phi_{L-2}}\circ \dots \circ g_{\phi_{1}} \circ g_{\phi_{0}}(\bz_0;\bPhi,\bR),\nonumber
\end{eqnarray}
where $\bUpsilon_2=\{\phi_i\}_{i=0}^{L-1}$ represents the learnable (tunable) parameters of the decoder function. Note that we have over-parameterized the iterations of the RFPI algorithm by introducing a new variable $\btau_i$ at each iteration for the sparsity inducing step (i.e., Eq. \eqref{eq:10b}). Moreover, in contrast with the original iterations, we have introduced a new step-size $\delta_i$ at each step of the iteration as well. Hence, the above decoder function can be interpreted as performing $L$ iterations of the original RFPI algorithm with an additional $L(n+1) - 2$ degrees of freedom (as compared to the base algorithm) expressed in terms of the set of the shrinkage thresholds $\btau_i$ and the gradient step-sizes $\delta_i$, i.e. $\{\btau_i, \delta_i\}_{i=0}^{L-1}$. Hence, the proposed decoder function is much more expressive than that of the iterations of RFPI algorithm.
\vspace{-4.2mm}
\subsection{Loss Function Characterization and Training}
The training of an AE should be carried out by defining a proper loss function $\calG\left(\bx,f^{\text{Decoder}}_{\bUpsilon_2} \circ f^{\text{Encoder}}_{\bUpsilon_1}(\bx)\right)$ that provides a measure of the similarity between the input and the output of the AE. The goal is to minimize the distance between the input target signal $\bx$ and the recovered signal $\hat{\bx}$ according to a similarity criterion. A widely-used option for the loss function is the output MSE loss, i.e., $\expecE_{\bx\sim\calD(\bx)}\left\{\|\bx - \hat{\bx}\|_2^2\right\}$.
Nevertheless, in deep architectures with a high number of layers and parameters, such a simple choice of the loss function makes it difficult to back-propagate the gradients, and hence, the vanishing gradient problem arises. Therefore, for the training of the proposed AE, a better choice for the loss function is to consider the cumulative MSE loss of all layers. As a result, one can also feed-forward the decoder function for only $l<L$ layers (a lower complexity decoding), and consider the output of the $l$-th layer as a good approximation of the target signal. For training, one needs to consider the constraint that the gradient step-sizes $\{\delta_i\}_{i=0}^{L-1}$ must be non-negative. By parameterizing the decoder function on the step-sizes and the shrinkage step thresholds, we need to regularize the training loss function ensuring that the network chooses positive step sizes and thresholds at each layer. With this in mind, we suggest the following loss function for training the proposed one-bit compressive AE. Let $\tilde{g}_i = g_{\phi_i} \circ g_{\phi_{i-1}} \circ \dots \circ  g_{\phi_{0}} \circ f_{\bUpsilon_1}^{\text{Encoder}}(\bx)$, and define the loss function for training as
\begin{align}
\calG(\bx;\hat{\bx}) = &\underbrace{\sum_{i=0}^{L-1}w_i ||\bx - \tilde{g}_i(\bx_{i})||_2^2}_{\text{accumulated MSE loss of all layers}} + \\
&\underbrace{\lambda\sum_{i=0}^{L-1} \relu(-[\bdelta]_i) + \lambda\sum_{i=0}^{nL-1} \relu(-[\btau]_i)}_{\text{regularization term for the step-sizes and shrinkage thresholds}},
\nonumber
\end{align}
where $\lambda>0$, $[\bdelta]_i = \delta_i$, and $\btau = [\btau_0^T, \dots,\btau_{L-1}^T]$.

	
	\section{Numerical Results}\label{sec:numerical}

In this section, we present the simulation results that investigate the performance of the proposed one-bit compressive~AE. For training purposes, we randomly generate $K$-sparse signals of length $n=128$, on the unit sphere, i.e. $\bx\in\realR^{128}$, and $\|\bx\|_2 = 1$. Furthermore, we fix the total number of layers of the decoder function to $L=30$; equivalent of performing only 30 optimization iterations of the form~\eqref{eq:10}. As for the sensing matrix to be learned, we assume $\bPhi\in\realR^{512\times 128}$. The results presented here are averaged over $128$ realizations of the system parameters. Although we consider the case that $m>n$, due to the focus of this study on one-bit sampling where usually a large number of one-bit samples are available, as opposed to the usual CS settings.\looseness=-1


The proposed one-bit CS AE is implemented using the $\mathrm{PyTorch}$ library \cite{paszke2017automatic}. The Adam algorithm \cite{kingma2014adam} with a learning rate of $10^{-3}$ is utilized for optimization of parameters of the proposed AE. We perform a batch-learning approach with mini-batches of size $64$, and a total number of $8000$ epochs. For training of the the proposed AE, we fix $K=16$, and evaluate the performance of the proposed method on target signals with $K=16$, as well as $K=24$ (which was not shown to the network during the training phase).  In all scenarios, the initial starting point of the optimization algorithms are the same. \looseness=-1


Fig.~\ref{fig:2} illustrates MSE for the recovered signal versus total number of optimization iterations $i$, for $i= 0,\dots,29$, and for sparsity levels  (a) $K=16$ and (b) $K=24$. We compare our algorithm with the RFPI iterations in \eqref{eq:16}-\eqref{eq:19}, in the following scenarios:\\ $\bullet$ {\emph{Case 1}}: The RFPI algorithm with a randomly generated sensing matrix whose elements are i.i.d and sampled from $\calN(0,1)$, and fixed values for $\delta$, and $\alpha$.\\ $\bullet$ {\emph{Case 2:}} The RFPI algorithm where the learned $\bPhi$ is utilized and the values for $\delta$ and $\alpha$ are fixed as the previous case.\\ $\bullet$ {\emph{Case 3:}} The RFPI algorithm with a randomly generated $\bPhi$ (same as Case 1), however, the learned shrinkage thresholds vector $\{\btau_i\}_{i=0}^{L-1}$ is utilized with a fixed step size.\looseness=-1\\ $\bullet$ {\emph{Case 4:}} The proposed one-bit CS AE method corresponding to the iterations of the form \eqref{eq:10a}-\eqref{eq:10d}, with the learned $\bPhi$, $\{\delta_i\}_{i=1}^{L-1}$, and $\{\btau_i\}_{i=0}^{L-1}$.

\vspace{-4.2mm}
\subsection{Discussion and Concluding Remarks}
It can be seen from Fig.~\ref{fig:2} that in both cases of $K \in \{16,24\}$, the proposed method demonstrates a significantly better performance than that of the RFPI algorithm (described in Case 1)---an improvement of $\sim 10$ times in MSE outcome. Furthermore, the effectiveness of the learned $\bPhi$ (Case 2), and the learned $\{\btau_i\}$ (Case 3) compared to the base algorithm (Case 1), are clearly evident, as both algorithms with learned parameters significantly outperform the original RFPI. Finally, although we trained the network for $K=16$ sparse signals, it still shows good generalization properties even for $K=24$ (see Fig.~\ref{fig:2} (b)). This is presumably due to the fact that the proposed AE is a hybrid model-based data-driven approach that exploits the existing domain knowledge of the problem as well as the available data at hand. Furthermore, note that the proposed method achieves high accuracy very quickly and does not require solving \eqref{eq:6} for several instances as opposed to the original RFPI algorithm---thus showing great potential for usage in real-time applications.


	\balance
	\bibliographystyle{IEEEbib}
	\bibliography{refs}

\begin{thebibliography}{10}

\bibitem{4472240}
E.~J. {Candes} and M.~B. {Wakin},
\newblock ``An introduction to compressive sampling,''
\newblock {\em IEEE Signal Processing Magazine}, vol. 25, no. 2, pp. 21--30,
  March 2008.

\bibitem{4558487}
P.~T. {Boufounos} and R.~G. {Baraniuk},
\newblock ``1-bit compressive sensing,''
\newblock in {\em 2008 42nd Annual Conference on Information Sciences and
  Systems}, March 2008, pp. 16--21.

\bibitem{5955138}
J.~N. {Laska}, Z.~{Wen}, W.~{Yin}, and R.~G. {Baraniuk},
\newblock ``Trust, but verify: Fast and accurate signal recovery from 1-bit
  compressive measurements,''
\newblock {\em IEEE Transactions on Signal Processing}, vol. 59, no. 11, pp.
  5289--5301, Nov 2011.

\bibitem{6418031}
L.~{Jacques}, J.~N. {Laska}, P.~T. {Boufounos}, and R.~G. {Baraniuk},
\newblock ``Robust 1-bit compressive sensing via binary stable embeddings of
  sparse vectors,''
\newblock {\em IEEE Transactions on Information Theory}, vol. 59, no. 4, pp.
  2082--2102, April 2013.

\bibitem{6404739}
A.~{Movahed}, A.~{Panahi}, and G.~{Durisi},
\newblock ``A robust {RFPI}-based 1-bit compressive sensing reconstruction
  algorithm,''
\newblock in {\em 2012 IEEE Information Theory Workshop}, Sep. 2012, pp.
  567--571.

\bibitem{6178284}
M.~{Yan}, Y.~{Yang}, and S.~{Osher},
\newblock ``Robust 1-bit compressive sensing using adaptive outlier pursuit,''
\newblock {\em IEEE Transactions on Signal Processing}, vol. 60, no. 7, pp.
  3868--3875, July 2012.

\bibitem{zhang2014efficient}
L.~Zhang, J.~Yi, and R.~Jin,
\newblock ``Efficient algorithms for robust one-bit compressive sensing,''
\newblock in {\em International Conference on Machine Learning}, 2014, pp.
  820--828.

\bibitem{khobahi2018signal}
S.~Khobahi and M.~Soltanalian,
\newblock ``Signal recovery from 1-bit quantized noisy samples via adaptive
  thresholding,''
\newblock in {\em 52nd Asilomar Conference on Signals, Systems, and Computers}.
  IEEE, 2018, pp. 1757--1761.

\bibitem{ameri2019one}
A.~Ameri, A.~Bose, J.~Li, and M.~Soltanalian,
\newblock ``One-bit radar processing with time-varying sampling thresholds,''
\newblock {\em IEEE Transactions on Signal Processing}, vol. 67, no. 20, pp.
  5297--5308, 2019.

\bibitem{roth2018comparison}
K.~Roth, H.~Pirzadeh, A.~Lee Swindlehurst, and J.~A. Nossek,
\newblock ``A comparison of hybrid beamforming and digital beamforming with
  low-resolution {ADCs} for multiple users and imperfect {CSI},''
\newblock {\em IEEE Journal of Selected Topics in Signal Processing}, vol. 12,
  no. 3, pp. 484--498, 2018.

\bibitem{zahabi2019one}
S.~J. Zahabi, M.~M. Naghsh, M.~Modarres-Hashemi, and J.~Li,
\newblock ``One-bit compressive radar sensing in the presence of clutter,''
\newblock {\em IEEE Transactions on Aerospace and Electronic Systems}, 2019.

\bibitem{Li2018}
Z.~Li, W.~Xu, X.~Zhang, and J.~Lin,
\newblock ``A survey on one-bit compressed sensing: theory and applications,''
\newblock {\em Frontiers of Computer Science}, vol. 12, no. 2, pp. 217--230,
  Apr 2018.

\bibitem{7780424}
K.~Kulkarni, S.~Lohit, P.~Turaga, R.~Kerviche, and A.~Ashok,
\newblock ``Reconnet: Non-iterative reconstruction of images from compressively
  sensed measurements,''
\newblock {\em arXiv preprint arXiv:1601.06892}, 2016.

\bibitem{ILIADIS20189}
M.~Iliadis, L.~Spinoulas, and A.~K. Katsaggelos,
\newblock ``Deep fully-connected networks for video compressive sensing,''
\newblock {\em Digital Signal Processing}, vol. 72, pp. 9 -- 18, 2018.

\bibitem{7447163}
A.~{Mousavi}, A.~B. {Patel}, and R.~G. {Baraniuk},
\newblock ``A deep learning approach to structured signal recovery,''
\newblock in {\em 2015 53rd Annual Allerton Conference on Communication,
  Control, and Computing (Allerton)}, Sep. 2015, pp. 1336--1343.

\bibitem{hershey2014deep}
J.~R. Hershey, J.~L. Roux, and F.~Weninger,
\newblock ``Deep unfolding: Model-based inspiration of novel deep
  architectures,''
\newblock {\em arXiv preprint arXiv:1409.2574}, 2014.

\bibitem{8683876}
S.~{Khobahi}, N.~{Naimipour}, M.~{Soltanalian}, and Y.~C. {Eldar},
\newblock ``Deep signal recovery with one-bit quantization,''
\newblock in {\em ICASSP 2019 - 2019 IEEE International Conference on
  Acoustics, Speech and Signal Processing (ICASSP)}, May 2019, pp. 2987--2991.

\bibitem{Plan2013}
Y.~Plan and R.~Vershynin,
\newblock ``One-bit compressed sensing by linear programming,''
\newblock {\em Communications on Pure and Applied Mathematics}, vol. 66, no. 8,
  pp. 1275--1297, 2013.

\bibitem{6638799}
Y.~{Shen}, J.~{Fang}, H.~{Li}, and Z.~{Chen},
\newblock ``A one-bit reweighted iterative algorithm for sparse signal
  recovery,''
\newblock in {\em 2013 IEEE International Conference on Acoustics, Speech and
  Signal Processing}, May 2013, pp. 5915--5919.

\bibitem{paszke2017automatic}
A.~Paszke, S.~Gross, S.~Chintala, G.~Chanan, E.~Yang, Z.~DeVito, Z.~Lin,
  A.~Desmaison, L.~Antiga, and A.~Lerer,
\newblock ``Automatic differentiation in pytorch,''
\newblock 2017.

\bibitem{kingma2014adam}
D.~P. Kingma and J.~Ba,
\newblock ``Adam: A method for stochastic optimization,''
\newblock {\em arXiv preprint arXiv:1412.6980}, 2014.

\end{thebibliography}

\end{document}